\documentclass[acmtog]{acmart}

\usepackage{booktabs} 

\usepackage[ruled]{algorithm2e} 

\usepackage{multirow}
\usepackage{adjustbox}
\usepackage{rotating}
\usepackage{graphicx}

\SetAlFnt{\small}
\SetAlCapFnt{\small}
\SetAlCapNameFnt{\small}
\SetAlCapHSkip{0pt}

\acmJournal{TOG}

\begin{document}
\title{End-to-End Adaptive Monte Carlo Denoising and Super-Resolution}

\author{Xinyue Wei}
\affiliation{%
 \institution{University of California San Diego}
 \country{USA}}
\affiliation{%
 \institution{Tencent AI Lab}
 \country{China}
}
\email{sarahwei0210@gmail.com}
\author{Haozhi Huang}
\affiliation{%
 \institution{Tencent AI Lab}
 \country{China}
}
\email{huanghz08@gmail.com}
\author{Yujin Shi}
\affiliation{%
 \institution{Tencent AI Lab}
 \country{China}
}
\email{shi.yujin@qq.com}
\author{Hongliang Yuan}
\affiliation{%
 \institution{Tencent AI Lab}
 \country{China}
}
\email{11488336@qq.com}
\author{Li Shen}
\affiliation{%
 \institution{Tencent AI Lab}
 \country{China}
}
\email{mathshenli@gmail.com}
\author{Jue Wang}
\affiliation{%
 \institution{Tencent AI Lab}
 \country{China}
}
\email{arphid@gmail.com}

\renewcommand\shortauthors{Wei, X. et al}

\begin{abstract}
\footnotetext{This work was done when Xinyue Wei was an intern at Tencent AI Lab.}

The classic Monte Carlo path tracing can achieve high quality rendering at the cost of heavy computation. Recent works make use of deep neural networks to accelerate this process, by improving either low-resolution or fewer-sample rendering with super-resolution or denoising neural networks in post-processing. However, denoising and super-resolution have only been considered separately in previous work. We show in this work that Monte Carlo path tracing can be further accelerated by joint super-resolution and denoising (SRD) in post-processing. This new type of joint filtering allows only a low-resolution and fewer-sample (thus noisy) image to be rendered by path tracing, which is then fed into a deep neural network to produce a high-resolution and clean image. 
The main contribution of this work is a new end-to-end network architecture, specifically designed for the SRD task. It contains two cascaded stages with shared components. We discover that denoising and super-resolution require very different receptive fields, a key insight that leads to the introduction of deformable convolution into the network design. Extensive experiments show that the proposed method outperforms previous methods and their variants adopted for the SRD task.


\end{abstract}

%


%
%


\begin{teaserfigure}
  \centering
  \includegraphics[width=\linewidth]{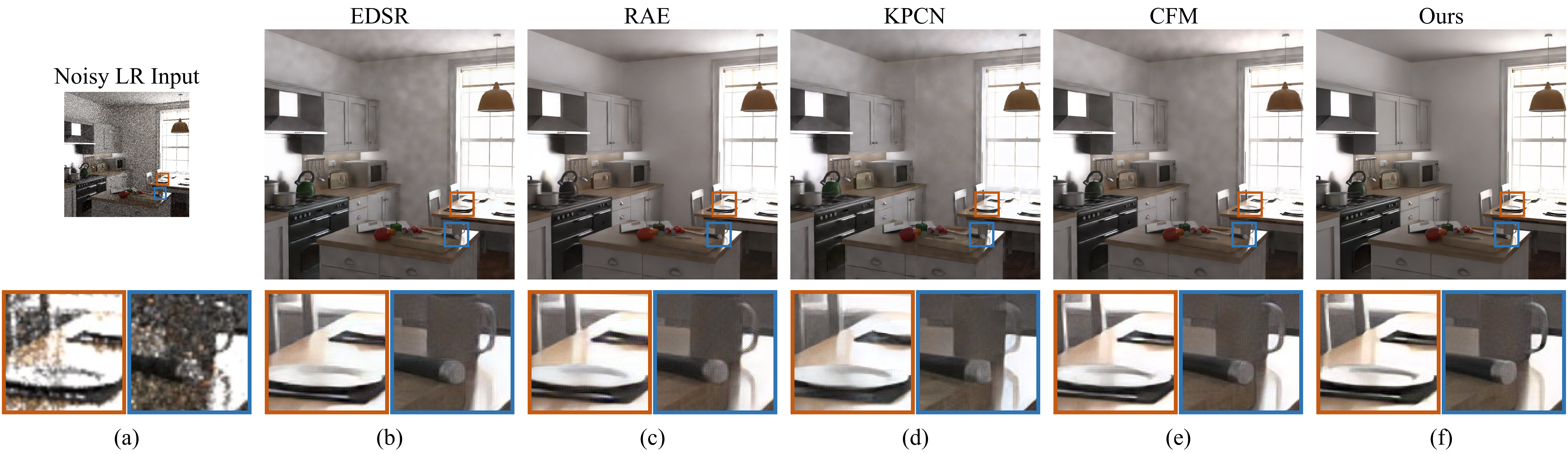}
    \vspace{-0.25in}
  \caption{Left to right: (a) low-resolution (540p) noisy (8spp) image generated by path tracing rendering; (b) EDSR~\cite{lim2017enhanced} (relMSE=0.0101); (c) RAE~\cite{chaitanya2017interactive} (relMSE=0.0115); (d) KPCN~\cite{bako2017kernel} (relMSE=0.0131); (e) CFM~\cite{xu2019adversarial} (relMSE=0.0094); (f) Our (relMSE=0.0068). We improve Monte Carlo rendering by exploring a new strategy: reducing the resolution and the sampling rate at the same time. All the outputs in (b)-(f) are 1080p. The reference image is generated using 2048spp.}
\end{teaserfigure}

\maketitle
\section{Introduction}



Path tracing is a widely-used rendering method in film production due to its favorable quality and stability. In recent years, modern video games start to use path tracing to achieve high visual quality rendering. This new trend puts real-time path tracing in great need, but meeting the increasingly higher demand for better quality and higher resolution is challenging due to the inherent computational complexity of path tracing. 

Recently, deep neural networks (DNNs) have shown great potential in many visual rendering tasks either for speeding up the pipeline, or improving the visual quality of the rendered results. Following these successes some have also tried to use neural networks to accelerate path-tracing-based rendering. Previous methods in this domain adopt two different methodologies: super-resolution and denoising. The former uses path tracing to render lower resolution imagery, then applies upsampling techniques such as DLSS~\cite{edelsten2019truly}, or temporal neural network by Lei \textit{et al.}~\cite{xiao2020neural} to increase its resolution while maintaining image sharpness. 
The latter renders imagery at the desired resolution directly but with fewer color samples per pixel, leading to noisy output. DNNs are then used to denoise the output to restore clean imagery.   
Specifically, pixel-based denoising methods~\cite{chaitanya2017interactive,bako2017kernel,vogels2018denoising,xu2019adversarial} either directly regress the final pixel colors or predict pixel-centric filters. In contrast, sample-based denoising methods~\cite{gharbi2019sample,munkberg2020neural} work in the sample space, filtering individual raw samples at each pixel location.

In this work, we aim to further improve the efficiency of path tracing by investigating a new strategy that has not been explored before: reducing the resolution and the sampling rate at the same time. This leads to a low-resolution, noisy initial result, which will then be processed by a novel neural network to perform joint denoising and super-resolution. The advantage of this approach is obvious: by reducing both resolution and sampling rate, the computational cost of the main bottleneck of the pipeline: Monte Carlo rendering, is greatly reduced. The heavy burden of producing a high quality result is then shifted onto the following neural network, which needs to excel in both super-resolution and denoising while maintaining high computational efficiency. We call this new task SRD (Super-Resolution and Denoising) and focus on providing such a neural network in this paper. Time evaluation in Sec.~\ref{sec:time_eval} shows that we can reduce the time-consuming of a 8spp rendering to almost one-fifth via reducing the resolution from 1080p to 540p.

A straightforward solution is to separately train two deep neural networks, one for super-resolution and the other for denoising, and apply them sequentially. We have tried this approach and found the results to be unsatisfactory.  The main reason is that the two networks trained in this way have no strong entanglement, and the second network often fails to produce good results by ignoring the information of the first network. Another approach is to re-train one of the current super-resolution or denoising networks for the SRD task, by changing the network input to low-resolution and noisy images during training. We also tried this approach and failed to produce high quality results, as we discovered that the network characteristics needed for super-resolution and denoising are fairly different, among which the receptive field is an important factor (see Sec~\ref{sec:rf_require}).
This important observation serves as a cornerstone for the proposed method. 

In this paper, we propose an end-to-end architecture specifically designed for the SRD task. Our model still has two stages, but in contrast to the direct concatenation of two separate networks, we propose a set of techniques to strengthen the entanglement of the two stages, such as shuffling features from the former model to the latter one and using intermediate supervision to reinforce the learning of each stage. Second, based on our discovery of different receptive field requirements of the two stages, we design different architectures for them. Furthermore, to avoid manually choosing the most appropriate receptive field size of each stage, 
we introduce \textit{deformable convolution} to our architecture, which has proven its effectiveness on object detection and semantic segmentation, with a strong ability to adapt to different geometric structures. By doing so, the network can dynamically learn the appropriate receptive fields in each stage for specific image content.
This design directly leads to better performance against other state-of-the-art methods as well as various baseline approaches, as we will show in Sec~\ref{sec:exp_sota}.

Our contributions can be summarized as follows:
\begin{itemize}
    \item We propose a new task called SRD-based Monte Carlo rendering, which further accelerates the Monte Carlo rendering pipeline by introducing neural network based joint super-resolution and denoising. 
    \item We propose the first network design for this task based on insightful analysis of the network requirements for super-resolution and denoising.
    \item We introduced deformable convolution to adaptively learn the appropriate receptive field.
    \item Extensive experiments demonstrate that the proposed method achieves state-of-the-art performance both quantitatively and qualitatively.
\end{itemize}





\section{Related Work}

\subsection{Super Resolution}
Image super-resolution is an ill-posed problem that tries to recover high resolution content from its low resolution version. Dong \textit{et al.}~\cite{dong2014learning} is the first to use neural network SRCNN to learn the mapping from LR to HR image. ESPCN~\cite{shi2016real} proposed a sub-pixel convolution
layer and achieved real-time speed on 1080p image. SRResNet~\cite{ledig2017photo} applied residual blocks to the SR task and EDSR~\cite{lim2017enhanced} removed the BN layers in residual blocks and expanded model size to further improve the performance. Besides, residual learning~\cite{kim2016accurate}, laplacian pyramid structure~\cite{lai2017deep,lai2018fast}, dense connection~\cite{tong2017image,wang2018esrgan}, attention mechanism~\cite{zhang2018image}, information distillation~\cite{hui2018fast} are also introduced to real-world image super-resolution problem.

Super-resolution in rendering is also called supersampling. Deep Learning Super-Sampling (DLSS)~\cite{edelsten2019truly} replaces the traditional heuristics algorithms with neural networks to upscale low-resolution renderings in real-time, whose implementation details are unavailable. Xiao \textit{et al.}~\cite{xiao2020neural} proposed to use temporal dynamics and gBuffer information to achieve improvement in the challenging 4$\times$4 upsampling case. Our work focuses on 2$\times$2 upscaling, coupled with the denoising task to explore the full potential of an end-to-end model.

\subsection{Monte Carlo Denoising}
A comprehensive review of Monte Carlo denoising is available in the work of Zwicker \textit{et al.}~\cite{zwicker2015recent}, and we only focus on the methods based on neural networks, which are the most relevant. Current neural Monte Carlo denoising methods can be divided into pixel-based and sample-based ones. In Monte Carlo path tracing, the color of one pixel is the average of multiple sample rays. Pixel-based ones work directly on pixel values while sample-based methods work on individual samples. 

On the one hand, RAE~\cite{chaitanya2017interactive} and KPCN~\cite{bako2017kernel} are two concurrent pioneer works introducing neural networks to Monte Carlo denoising. The former designed a recurrent auto-encoder to interactively process input image of low sample rate; the latter made the network predict pixel-centric kernels for filtering, aiming at input images of higher sample rate.  Based on KPCN, Vogels \textit{et al.}~\cite{vogels2018denoising} expanded kernel-prediction models with task-specific modules and an asymmetric loss. Later on, Xu \textit{et al.}~\cite{xu2019adversarial} introduced a conditional module to better utilize auxiliary features and generative adversarial networks to achieve more realistic results. In this paper, we introduce perceptual loss~\cite{johnson2016perceptual}, which also plays the role of enhancing image perceptual quality but requires no extra training. 

On the other hand, the first sample-based method~\cite{gharbi2019sample} is proposed to handle defocus blur and depth of field but comes with high computational costs, especially when the sample rate is high. Munkberg \textit{et al.}~\cite{munkberg2020neural} introduced layer embedding strategy to divide samples into groups for reducing computational cost. From the perspective of sampling, Hasselgren \textit{et al.}~\cite{hasselgren2020neural} proposed adaptive sampling and optimized sampling and denoising jointly for better temporal stability.

In order to handle super-resolution and denoising simultaneously, we use pixel-based methods as our baseline and comparison, while sample-based methods are computationally heavy and cannot be practically adapted to the new task. 

\subsection{Adaptive Receptive Field}
People have explored ways to adapt the receptive field of neural convolutions. Atrous convolution~\cite{chen2017deeplab} explicitly increased the distance between filter samples to handle larger context without adding more parameters. Deformable convolution \textit{et al.} ~\cite{dai2017deformable,zhu2019deformable, gao2019deformable} was proposed to automatically learn the location of samples of a filter to adaptively change the receptive of field. These methods showed good improvement in object detection and semantic segmentation. While we introduce deformable convolution to our end-to-end architecture to meet the content-specific requirements of receptive field on the novel super-resolution and denoising task.
\section{Method}

\subsection{Problem Statement}
Instead of dealing with super-resolution or denoising individually like in previous works, we define a new task \textit{joint super-resolution and denoising (SRD)}, whose goal is to predict a high-resolution, noise-free result from a low-resolution and noisy rendered image.

In Monte Carlo path tracing, the radiance $c$ of a pixel can be approximated by the average of $k$ samples per pixel (\textit{k spp}). The more samples in each pixel, the less noise we have, but the more computation time it needs. To alleviate this problem, deep-learning-based MC denoising methods are proposed to learn the mapping from the low-spp input to the high-spp noise-free image, where the input consists of the noisy RGB image and the corresponding gBuffer data (albedo, normal, roughness, \textit{etc}). On the other hand, general super-resolution aims to learn the mapping from a low-resolution image $I^{LR}$ to a high-resolution image $I^{HR}$.


In SRD, our goal is to train a deep neural network $F$ to learn the mapping from a low-resolution and low-spp rendered result $c_p^{LR}$ to a high-resolution and high-spp one $\hat{c_p}^{HR}$:
\begin{equation}\label{func:prob_state}
     \hat{c_p}^{HR} = F(c_p^{LR}, g_p^{LR};\theta),
\end{equation}
where $g_p^{LR}$ represents its corresponding gBuffer data directly acquired from a path tracing renderer, $\theta$ is the parameters of the neural network $F$. Our optimization target is to minimize the difference between the predicted result $\hat{c_p}^{HR}$ and the high-resolution and high-spp ground truth $y_p^{HR}$: 
\begin{equation}
    \hat{\theta} = {\arg\min}_{\theta}{L(\hat{c_p}^{HR}, y_p^{HR})}
\end{equation}
where $L$ is the loss function evaluating the difference between the two images, which will be further explained in Sec.~\ref{sec:loss_func}.

\subsection{End-to-End Architecture Overview}\label{sec:overview}
Our end-to-end architecture contains two entangled stages: the super-resolution stage and the denoising stage. Our overall strategy is to handle the two tasks using different network structures but combine them in an end-to-end manner. An end-to-end model is better than concatenating two pretrained models because it strengthens the entanglement between the two stages and reinforces the information propagation. However, different structures are employed for different stages because the network characteristics needed for the two tasks are quite different, as we will demonstrate later.

\begin{figure*}[t]
\begin{center}
   \includegraphics[width=\linewidth]{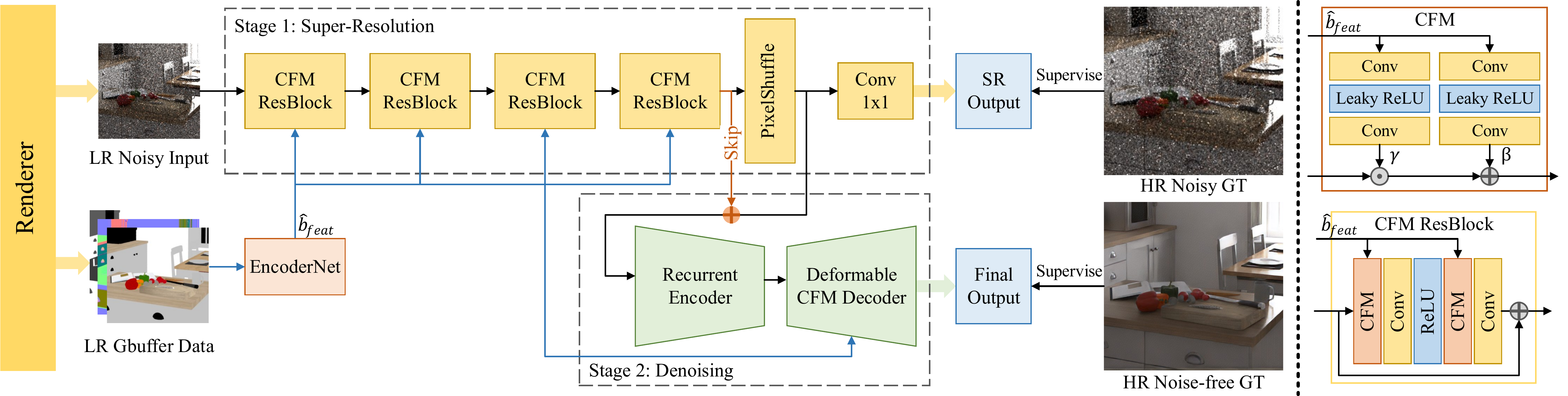}
\end{center}
    \vspace{-0.15in}
  \caption{End-to-End architecture overview. The whole network is divided into a super-resolution stage and a denoising stage. The first stage has four stacked residual blocks with 3$\times$3 convolution layers; the second stage is a deformable recurrent auto-encoder. Each stage has its corresponding supervision. Instead of being concatenated with the RGB input, our gBuffer information is inserted to the two stages through the CFM module. The right two blocks show how CFM works, that gBuffer data are used to normalize the input feature map.}
    \vspace{-0.1in}
\label{fig:pipeline}
\end{figure*}

We set super-resolution as the first stage and denoising the second one. This is under the consideration that super-resolution is for enhancing subtle image details, thus it should have access to the original, unfiltered signal. Our experimental results in Sec.~\ref{sec:exp_rf} show that reversing the order of the two stages would lead to lower quality output.
The complete pipeline is shown in Fig.~\ref{fig:pipeline}. At the first stage, the SR network upscales the noisy low-resolution image $c_p^{LR}$ into a preliminary high-resolution image. Then, the denoising network eliminates the noise as well as artifacts introduced by the SR stage. The final result is the high-resolution, noise-free image $\hat{c_p}^{HR}$. The noisy HR image is used as an intermediate supervision signal for the first stage. This extra supervision ensures an explicit boundary between the SR and denoising stages, and allows each stage to focus on its own task for better performance. 

In more detail, we apply stacked residual blocks with a pix-shuffle layer as the first stage network, which was proven to be effective in classic SR methods. Deformable recurrent auto-encoder is used as the second stage network. As for utilizing gBuffer data, we apply the conditioned feature modulation (CFM) proposed in ~\cite{xu2019adversarial} to normalize feature maps in each stage. The details of the architecture design and the deformable recurrent auto-encoder will be explained in Sec.~\ref{sec:info_reserve} and Sec.~\ref{sec:deform_conv}.

\subsection{Receptive Field Requirement of SR and Denoising}\label{sec:rf_require}
To design a network architecture that is suitable for handling both super-resolution and denoising, we explore the network characteristics needed by these two tasks. 

We start by adapting the classic super-resolution network EDSR~\cite{lim2017enhanced} and the denoising network RAE~\cite{chaitanya2017interactive} to solve both tasks, respectively. As expected, we found that EDSR performs better on the SR task, while RAE performs better on the denoising task. One of the major differences between EDSR and RAE is the receptive field. We thus assume that the preferred receptive fields for the two tasks are significantly different. To justify the assumption, we adopt atrous convolution~\cite{chen2017deeplab} to manually control the receptive field and further examine the performance difference. Specifically, we stack several residual blocks with atrous convolution, change the dilation rate to adjust the receptive field of the whole network. According to the results shown in Sec.~\ref{sec:exp_rf}, we conclude that denoising requires a much larger receptive field than super-resolution. The intuition behind this observation is that a large receptive field defines large neighborhoods, which are statistically necessary when denoising a heavy-corrupted image. Super-resolution, on the other hand, aims at recovering high frequency details that have very limited spatial correlations.

\subsection{Deformable Recurrent Auto-encoder}\label{sec:deform_conv}
\begin{figure}[t]
\begin{center}
   \includegraphics[width=\linewidth]{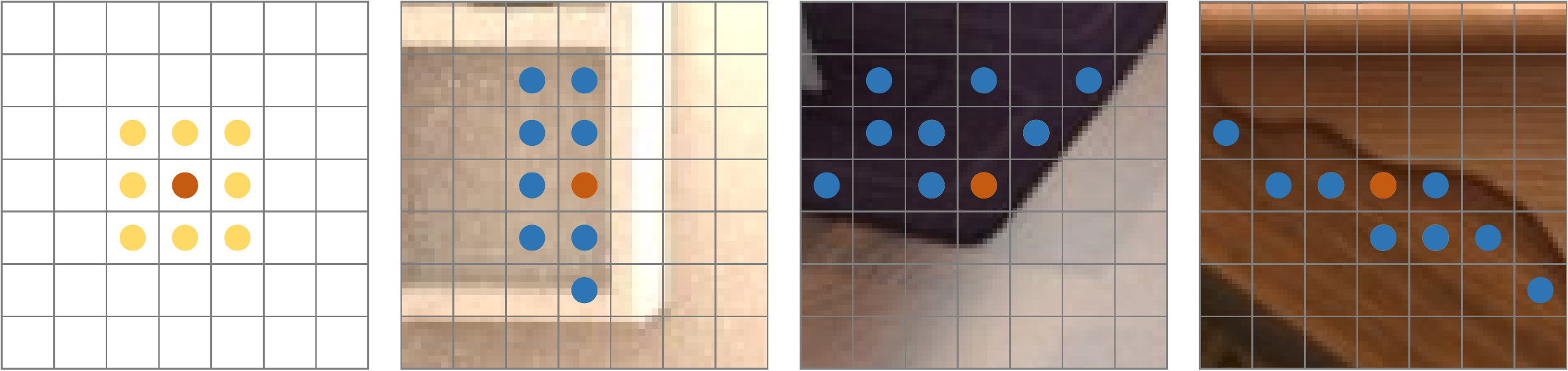}
\end{center}
    \vspace{-0.15in}
  \caption{Comparison of normal convolution samples (yellow points) and the ideal samples (blue points). Normal convolution always has an unchanged square receptive field while we prefer it to change the sampling region in terms of semantics with the help of deformable convolution.}
    \vspace{-0.1in}
\label{fig:deform_sample}
\end{figure}

Based on the analysis above, we seek to use different receptive fields in the two stages. Manually setting the appropriate receptive field is however not desirable. Moreover, this size of the receptive field is also scene-related. For example, as is shown in Fig.~\ref{fig:deform_sample}, for a pixel near a long edge, a stripe-shaped receptive field can better cover its semantic neighborhood. For a pixel near the corner, a fan-shaped receptive field is preferred. Previous learning-based MC denoising or super-resolution methods use traditional convolution layers to build the network, which has fixed square receptive fields once the architecture is determined. We thus aim to find an architecture that can automatically adjust the receptive field according to the input image content.

\begin{figure}[t]
\begin{center}
   \includegraphics[width=\linewidth]{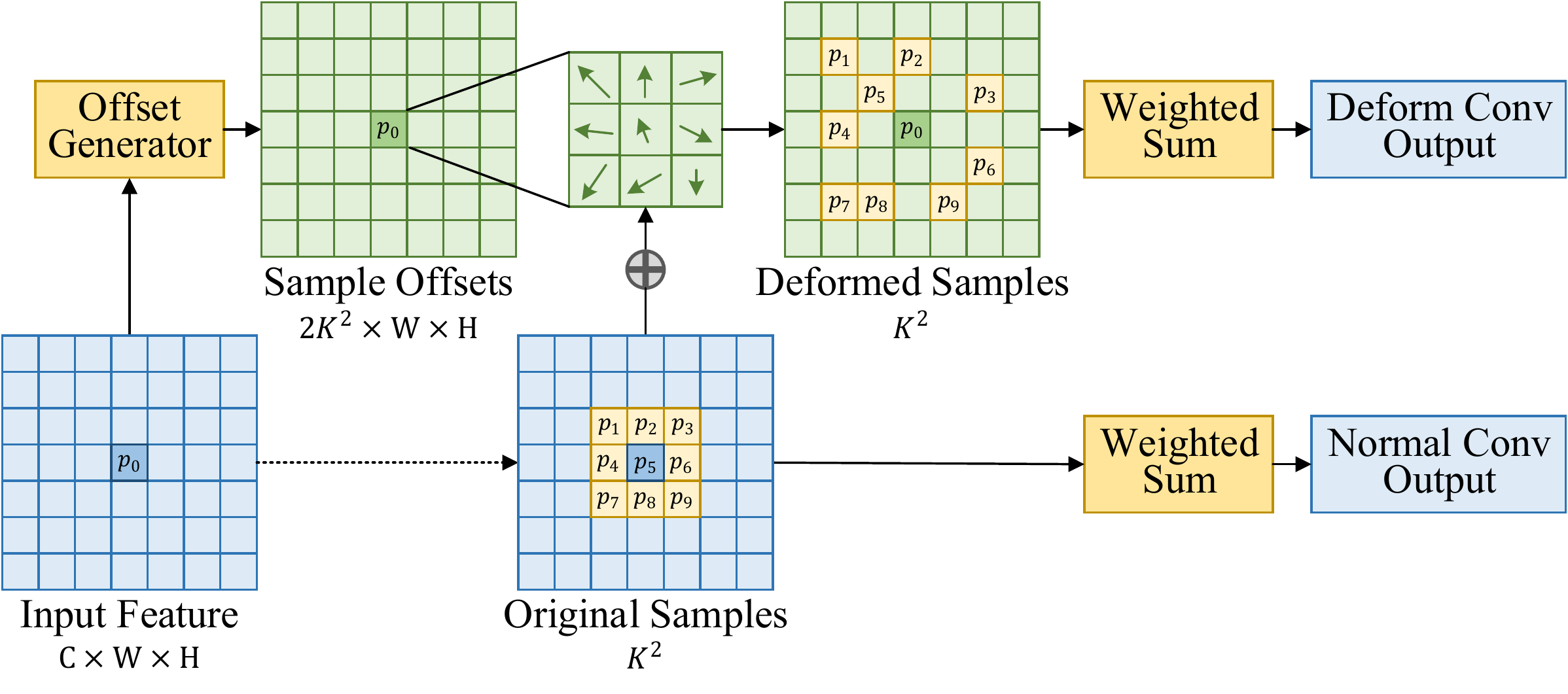}
\end{center}
    \vspace{-0.15in}
  \caption{Architecture of deformable convolution. Compared with normal convolution, deformable convolution additionally learns an offset for each sampling location, thus the actual sampling region is adaptively changing according to the input features.}
    \vspace{-0.1in}
\label{fig:deform_conv}
\end{figure}

Inspired by the success of deformable convolution in object detection and semantic segmentation~\cite{dai2017deformable,zhu2019deformable}, we propose to integrate deformable convolution into our architecture for the SRD task. A traditional convolution can be regarded as the weighted sum of samples within a fixed square, denoted as:
\begin{equation}
    y(p_0) = \sum_{p_n \in S}w(p_n)x(p_0+p_n)
\end{equation}
where $p_0$ is the target pixel location, $x$ denotes the input feature map and $S$ represents the square sample region. As is shown in Fig.~\ref{fig:deform_conv}, deformable convolution additionally learn an offset for each sample location, which can be denoted as:
\begin{equation}
    y(p_0) = \sum_{p_n \in S, \Delta p_n \in O}w(p_n)x(p_0+p_n+\Delta p_n)
\end{equation}
where $O = G(x)$ is the offset set predicted by the \textbf{Offset Generator} given the current feature map $x$, $\Delta p_n \in O$ is the corresponding offset for each sample location. For 2D convolution, $O$ is a $2K^2$-dim vector with offsets on both $x$ and $y$ directions, where $K$ refers to the kernel size of convolution. During implementation, offset generator is a 1$\times$1 convolution layer.

\begin{figure}[t]
\begin{center}
   \includegraphics[width=\linewidth]{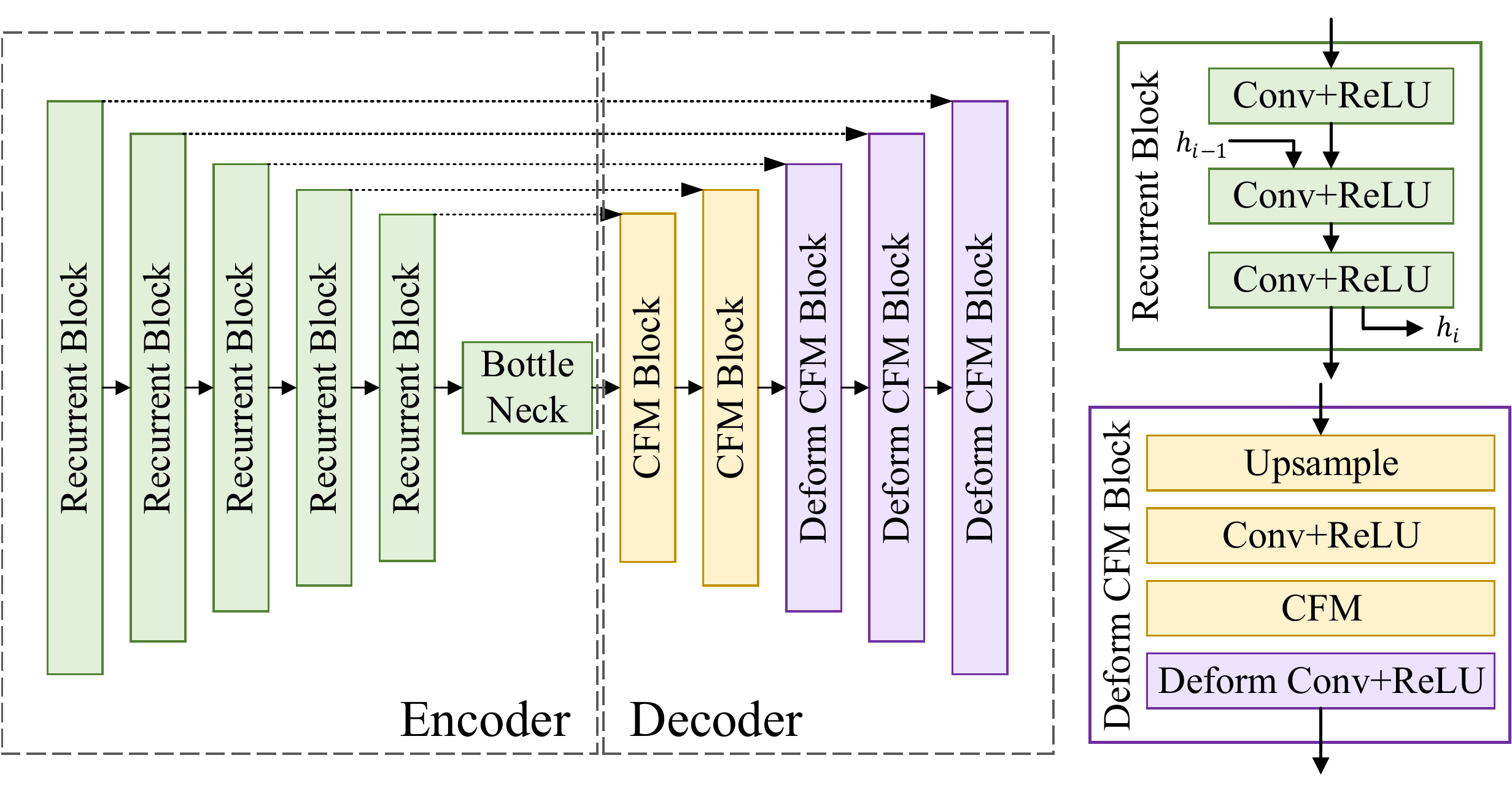}
\end{center}
    \vspace{-0.15in}
  \caption{Deformable recurrent auto-encoder details. 2$\times$2 maxpooling is adopted for the encoder, while nearest neighbor upsample is employed for the decoder. CFM is applied only in decoder to save calculation amount. For deformable CFM block, we replace the second convolution layer with deformable convolution in the CFM block.}
    \vspace{-0.1in}
\label{fig:rae}
\end{figure}

The architecture of deformable recurrent auto-encoder is shown in Fig.~\ref{fig:rae}, where the last three blocks in the decoder are the Deformable CFM blocks. We find that when using deformable convolution, there is a trade-off between model capacity and training complexity, as shown in Sec.~\ref{sec:deform_num}. Thus in our proposed architecture, we use three deformable blocks only in the denoising stage instead of both stages. For the SR stage, we experimentally find that the learned receptive field of the deformable block is similar to the normal convolution block, as shown in Sec.~\ref{sec:exp_deform_conv}. Thus, we do not apply deformable block in the SR stage.

\subsection{Information Propagation}\label{sec:info_reserve}
For an end-to-end architecture that contains two cascaded stages, it is important to keep valuable information propagating to the latter layers and to strengthen the entanglement between the two stages. We propose several techniques to better help information propagation.

Firstly, we increase the channel number of the last layer in the first stage. The original three-channel output (HR noisy) contains very little information for the second stage and is inadequate for end-to-end training.

We then design a skip connection between the two stages, shown as the red line in Fig.~\ref{fig:pipeline}. By element-wise adding the deep features before upsampling layer to the features of the same size in the later stage, denoising can reuse the deep features from the former stage directly.

Finally, we discard the traditional usage of the gBuffer data. In general MC denoising methods, high resolution gBuffer data is provided, while in SRD, we can only acquire low-resolution gBuffer data. Simply concatenating the gBuffer data with the rendered RGB image as input limits its effectiveness in our two-stage backbone, as the valuable information from the gBuffer data in one task is different from that in the other. To let the gBuffer data play the best role in both stages, we employ CFM to modulate the feature maps in each stage individually conditioning on the gBuffer data.

\subsection{Loss Functions}\label{sec:loss_func}
Similar to ~\cite{chaitanya2017interactive}, we utilize spatial loss, edge loss and temporal loss. Additionally, we innovatively introduce perceptual loss~\cite{johnson2016perceptual}. In Sec.~\ref{sec:exp_preceptual}, we demonstrate its effectiveness in better keeping details and reconstructing structures.

We use $L_1$ loss as the spatial loss, as it is widely used to evaluate the pixel-wise difference between two images, denoted as:
\begin{equation}
    L_s = \frac{1}{N}\sum_i^N{|\hat{y_i}-y_i|},
\end{equation}
where $N$ refers to the pixel number in an image, $\hat{y_i}=f(x_i)$ represents the predicted color, and $y_i$ is the ground truth color.

In order to strengthen the constraint to edges, we use gradient-domain $L_1$ loss as the edge loss, denoted as:
\begin{equation}
    L_g = \frac{1}{N}\sum_i^N{|\nabla{\hat{y_i}-\nabla{y_i}}|},
\end{equation}
where the gradient $\nabla{(\cdot)}$ is calculated by High Frequency Error Norm (HFEN)~\cite{ravishankar2010mr}. HFEN uses Laplacian filter to detect edges of a smoothed image by Gaussian filter. We use the same setting as in ~\cite{chaitanya2017interactive}, with the Gaussian kernel size $\sigma = 1.5$.

The temporal $L_1$ loss is to enhance temporal consistency between consecutive frames, denoted as:
\begin{equation}
    L_t = \frac{1}{N}\sum_i^N{|\frac{\partial{\hat{y_i}}}{\partial{t}}-\frac{\partial{y_i}}{\partial{t}}|}.
\end{equation}

For only using $L_1$ loss as the spatial content constraint can lead to blurry results~\cite{goodfellow2020generative,lotter2015unsupervised}, we introduce perceptual loss to encourage the predicted result to retain more details and structural information from the ground truth. Perceptual loss~\cite{johnson2016perceptual} is designed to evaluate the \textit{perceptual similarity} between two images by comparing their high-level features from an ImageNet-pretrained VGG16~\cite{simonyan2014very} network. The loss function can be written as:
\begin{equation}
    L_p^{\phi,j} = \frac{1}{C_j H_j W_j}||\phi_j(\hat{y})-\phi_j(y)||_2^2
\end{equation}
where $\phi$ is the VGG-16 network, $\phi_j(x)$ denotes the $j$-th feature map in the shape of $C_j \times H_j \times W_j$. In practice, we calculate the loss at layer $relu3\_3$.

In summary, our final loss is a weighted sum of four terms, denoted as:
\begin{equation}
    L = w_g L_g+w_t L_t+w_s L_s+w_p L_p
\end{equation}
where $w_g, w_t, w_s, w_p$ are the weights for each loss terms. For the intermediate supervision, the ratio is set as $1:1:8:0$ as in ~\cite{chaitanya2017interactive}, which means the perceptual loss is inactive. For the final output supervision, the ratio is empirically set as $1:1:8:10$ after a simple grid search.


\section{Experimental Setup}
\subsection{Data}
\begin{figure}[t]
\begin{center}
   \includegraphics[width=\linewidth]{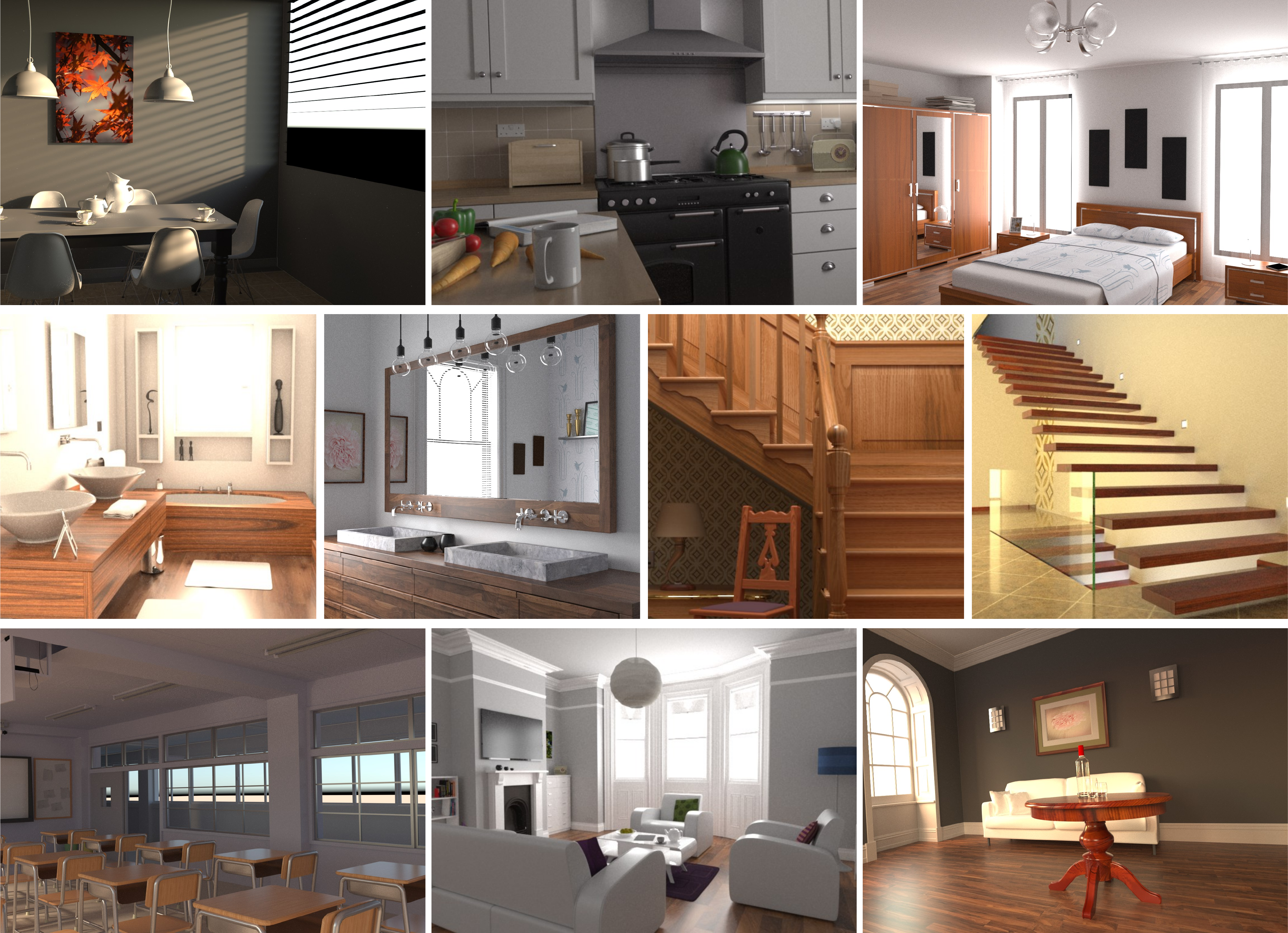}
\end{center}
    \vspace{-0.15in}
  \caption{Examples of 10 indoor scenes rendered by the Mitsuba path tracer. Cameras are controlled to collect data from various viewpoints for higher data diversity.}
\label{fig:scene}
\end{figure}

Deep learning methods require a large amount of training data to avoid over-fitting. We generate a dataset of 5,000 images with the Mitsuba path tracing renderer~\cite{Mitsuba}, including 4,000 for training and 1,000 for validation. We use 10 indoor scenes publicly available in ~\cite{bitterli2016rendering},  changing the camera viewpoint uniformly in the scene to increase data diversity, as shown in Fig.~\ref{fig:scene}. To evaluate the generalization ability of models, we create an additional \textbf{genearlization test set} including objects never used in the training data. For each training pair, the input image is at 8spp with the resolution of 960 $\times$ 540; while the reference image is at 2048spp with the resolution of 1920 $\times$ 1080.

During training, we randomly crop the images into 256 $\times$ 256 patches and normalize them into the range of $[0.0, 1.0]$. We only use three types of gBuffer data, including albedo, normal and roughness, to save storage. And we don't separate diffuse and specular like some works~\cite{bako2017kernel,xu2019adversarial} in order to save inference time.

\subsection{Implementation details}\label{sec:imp_detail}
We use PyTorch to implement the method and train the network on a single NVIDIA V100 GPU. In the end-to-end architecture, we simplify the original EDSR to contains 4 blocks and 32 channels in each block as the SR stage, with the output channel increased to 32. For the deformable recurrent auto-encoder, we adjust the channel of each convolution to 32. All the parameters in the network are initialized following He \textit{et al.}~\cite{he2015delving}.

During training, we make the input as a sequence of 7 frames to train the recurrent module. We train the network for 400 epochs using Adam optimizer~\cite{kingma2014adam}, with the learning rate 0.0006 and decay rates $\beta_1 = 0.9$ and $\beta_2 = 0.99$. The learning rate decays $50\%$ every 100 epochs. The batch size is set to 4 (sequences).

\subsection{Metrics and Comparisons}
We choose relative MSE~\cite{rousselle2011adaptive}, PSNR and SSIM~\cite{wang2004image} to evaluate the model performance.
To conduct a fair comparison, we adapt the super-resolution or denoising settings of the state-of-the-art methods to our SRD settings. For SR models, we change the training data from noise-free LR and HR image pairs into noisy LR and noise-free HR image pairs; for denoising models, we attach a pixel-shuffle layer~\cite{shi2016real} at the end of the original model for upscaling the result. We selected one super-resolution model EDSR~\cite{lim2017enhanced} and three denoising models RAE~\cite{chaitanya2017interactive}, KPCN~\cite{bako2017kernel} and CFM~\cite{xu2019adversarial} for comparison. We apply the same loss functions and training skills for all methods. In order to achieve a fair comparison, we restrict the floating-point operations per second (FLOPS) of each model to be around 260G Mac. The detailed numbers of FLOPS can be seen in Table~\ref{tab:sota}. The principle to modify the original models is adjusting the channel number without changing the typologies.
\section{Results}

\subsection{End-to-End Architecture Design}\label{sec:exp_rf}
\textbf{Experiments on Receptive Field.} In order to combine the SR and denoising tasks properly, we study the network characteristics needed by each task. As is stated in Sec.~\ref{sec:rf_require}, we design two experiments to study the different requirements of the two tasks.

Firstly, we adapt classic super-resolution method EDSR~\cite{lim2017enhanced} and pioneering learning-based MC denoising method RAE~\cite{chaitanya2017interactive} to accomplish the two tasks respectively. As shown in Table~\ref{tab:sr_denoise}, EDSR is good at super-resolution while RAE performs better on the denoising task. This is because the RAE model contains a larger receptive field thanks to the encoder-decoder structure.

\begin{table}
    \caption{\label{tab:sr_denoise}The performance of classic SR or denoising models.}
    \vspace{-0.15in}
\begin{center}
\resizebox{\linewidth}{!}{
\begin{tabular}{l|ccc|ccc}
\toprule
    \multirow{2}{*}{Model} & \multicolumn{3}{c|}{SR} & \multicolumn{3}{c}{Denoising} \\
    \cline{2-7}
    & relMSE$\downarrow$ & PSNR$\uparrow$ & SSIM$\uparrow$ & relMSE$\downarrow$ & PSNR$\uparrow$ & SSIM$\uparrow$  \\
\hline
EDSR & \textbf{0.0027} & \textbf{35.5950} & \textbf{0.9018} & 0.0051 & 32.1002 & 0.8671 \\
RAE & 0.0041 & 34.4967 & 0.8915 & \textbf{0.0031} & \textbf{34.3225} & \textbf{0.8800} \\
\bottomrule
\end{tabular}}
    \vspace{-0.1in}
\end{center}
\end{table}

\begin{figure}[t]
\begin{center}
   \includegraphics[width=\linewidth]{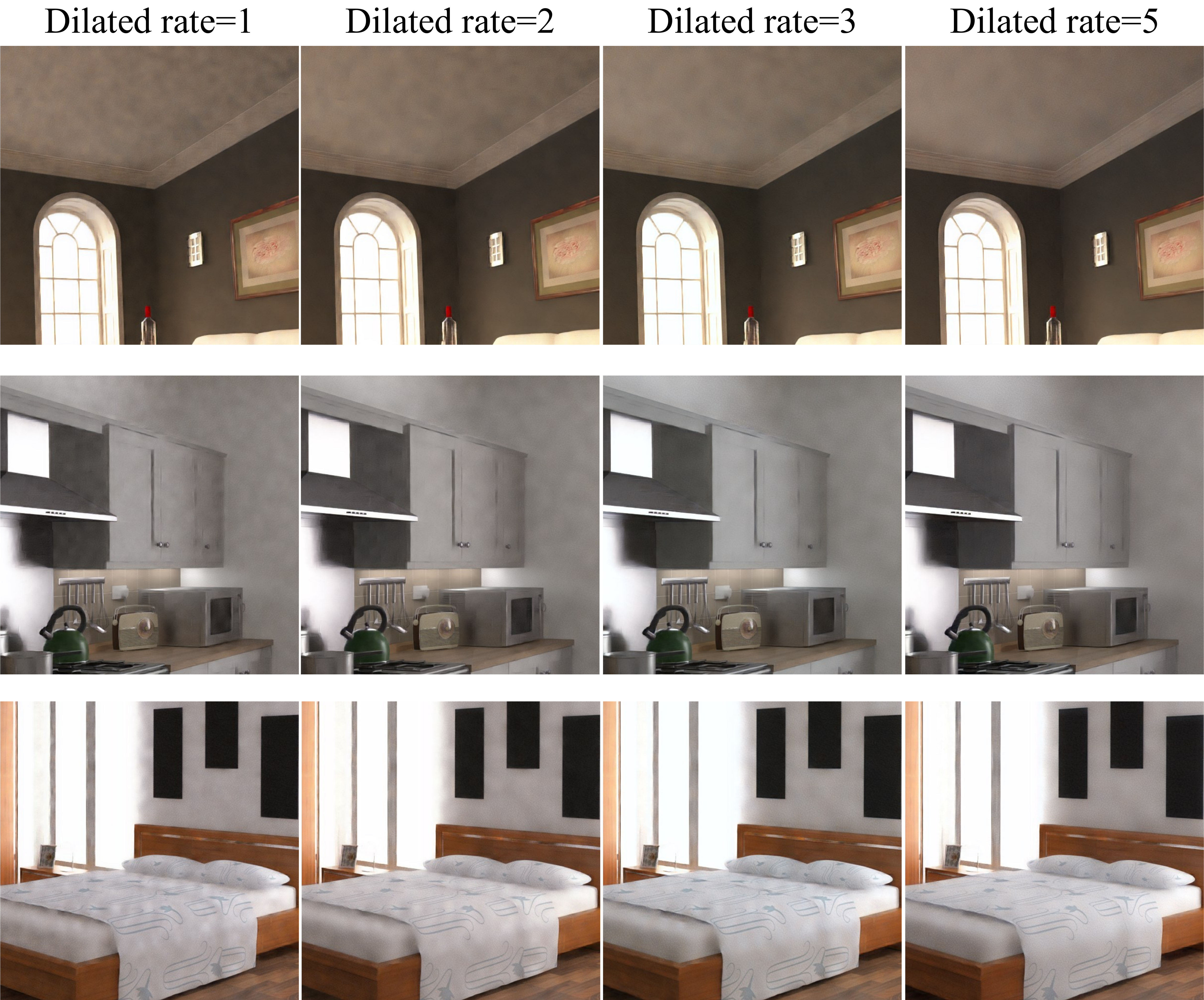}
\end{center}
    \vspace{-0.15in}
  \caption{Qualitative results of the denoising task when adjusting the model receptive field. The color mottle artifact in the output is less severe when the receptive field is larger.}
    \vspace{-0.1in}
\label{fig:dilated}
\end{figure}

Then we use atrous convolution to manually control the receptive field and study the performance change on SR or denoising respectively. We use EDSR as the backbone and replace the second convolution in each block by atrous convolution. By changing the dilation rate, the receptive field changes correspondingly. Table~\ref{tab:dilated} shows the results. As the receptive field increases, model performance on SR drops, while acting in the opposite way on the denoising task. The qualitative results of the denoising task better demonstrate the importance of the receptive field. As shown in Fig.~\ref{fig:dilated}, larger receptive fields lead to smoother results, 

In summary, we can conclude that super-resolution requires a smaller receptive field, while denoising requires a larger one, which is consistent with our assumption in Sec.~\ref{sec:rf_require}.

\begin{table}
    \caption{\label{tab:dilated}Evaluation on SR and denoising tasks when manually adjusting the receptive fields (RF) by changing the dilation rate. When receptive field increases, the performance on the SR task decreases. On the contrary, denoising prefers a larger receptive field.}
    \vspace{-0.15in}
\begin{center}
\resizebox{\linewidth}{!}{
\begin{tabular}{l|c|ccc|ccc}
\toprule
    \multirow{2}{*}{rate} & \multirow{2}{*}{RF} & \multicolumn{3}{c|}{SR} & \multicolumn{3}{c}{Denoising} \\
    \cline{3-8}
    & & relMSE$\downarrow$ & PSNR$\uparrow$ & SSIM$\uparrow$ & relMSE$\downarrow$ & PSNR$\uparrow$ & SSIM$\uparrow$  \\
\hline
1 & 25 & \textbf{0.0098} & \textbf{33.3536} & \textbf{0.8875} & 0.0071 & 30.3616 & 0.8437 \\
2 & 33 & 0.0115 & 32.6449 & 0.8816 & 0.0070 & 30.3792 & 0.8502 \\
3 & 41 & 0.0131 & 32.4612 & 0.8826 & 0.0050 & 31.9023 & 0.8497 \\
5 & 57 & 0.0131 & 32.6163 & 0.8832 & \textbf{0.0049} & \textbf{32.1816} & \textbf{0.8535} \\
\bottomrule
\end{tabular}}
    \vspace{-0.1in}
\end{center}
\end{table}

\noindent\textbf{Stage Combination and Order.} Known the different receptive field requirements of the two tasks, we design an end-to-end network with two stages coupled together to achieve better performance. The way and order of combining the two stages are also important factors in the final architecture design.

\begin{figure*}[t]
\begin{center}
   \includegraphics[width=\linewidth]{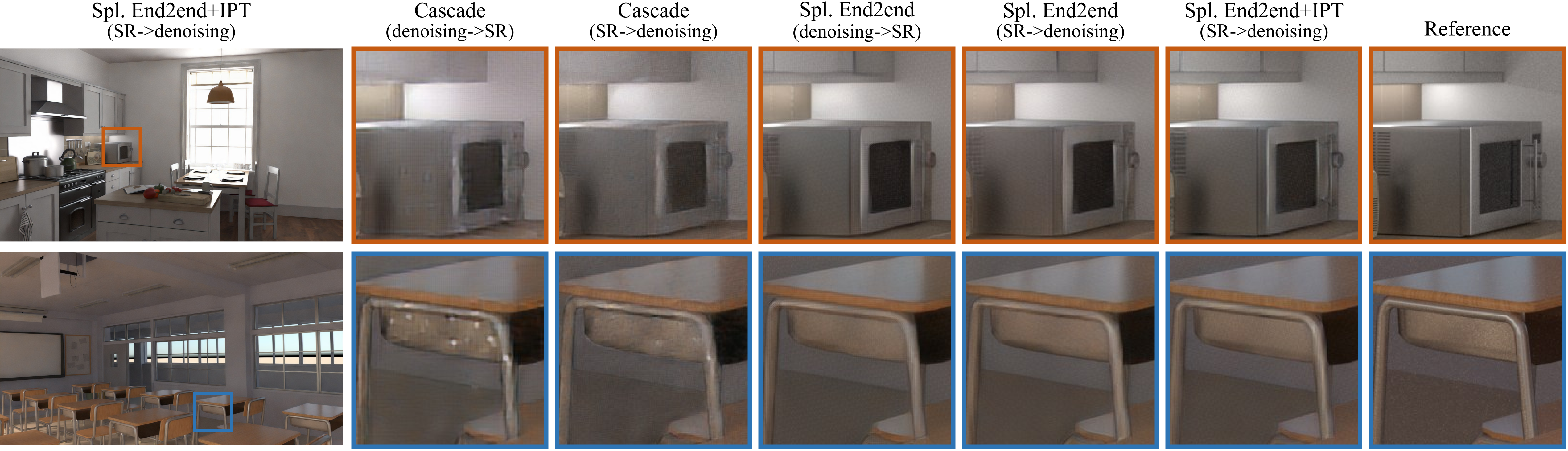}
\end{center}
    \vspace{-0.15in}
  \caption{Qualitative results of different variants when exploring the design of our end-to-end architecture. We compare different connecting ways and orders of the two-stage networks. \textit{IPT} refers to our proposed information propagation techniques. Obvious difference are around the microwave oven handle and the drawer edges.}
    \vspace{-0.1in}
\label{fig:end2end}
\end{figure*}

We start by cascading two pre-trained models (\textit{Cascade}). Then, we simply connecting two models as an end-to-end network (\textit{Simple End2End}). Finally, we design a bunch of information propagation techniques (IPT) for a better end-to-end architecture, as introduced in Sec.~\ref{sec:info_reserve}. For both \textit{Cascade} and \textit{Simple End2End}, we design two versions, doing super-resolution first (SR->denoising) or doing denoising first (denoising->SR). Results in Table~\ref{tab:arch_design} shows that doing SR first always has better performance, which is consistent with our intuition in Sec.~\ref{sec:overview}. According to the qualitative results in Fig~\ref{fig:end2end}, we found that conducting denoising first leads to extra noises while doing SR first achieves smoother results. Compared with \textit{Cascade}, \textit{Simple End2End} does not have obvious noise or overblur. After adding IPT to enforce information propagation, we acquire the final end-to-end model, which makes the results smoother and details clearer. The ablation study about the effectiveness of each IPT technique will be further explained in Sec.~\ref{sec:exp_info_reserve}.

Apart from overall architecture improvement for the SR and denoising task, we also explore a way to adaptively learn the appropriate receptive field, which will be explained in the next section.

\begin{table}
    \caption{\label{tab:arch_design}Comparison between different pre-trained and end-to-end solutions. \textit{Cascade} refers to connecting two pretrained models while \textit{Simple End2End} refers to training a two-stage network in an end-to-end manner. \textit{IPT} refers to our proposed information propagation techniques.}
    \vspace{-0.15in}
\begin{center}
\resizebox{\linewidth}{!}{
\begin{tabular}{l|ccc}
\toprule
Model & relMSE$\downarrow$ & PSNR$\uparrow$ & SSIM$\uparrow$ \\
\hline
Cascade (denoising->SR) & 0.0206 & 27.1596 & 0.8334 \\
Cascade (SR->denoising) & 0.0176 & 28.7736 & 0.8195 \\
Simple End2End (denoising->SR) & 0.0097 & 30.3543 & 0.8349 \\
Simple End2End (SR->denoising) & 0.0094 & 30.4851 & 0.8432 \\
Simple End2End (SR->denoising)+IPT & \textbf{0.0076} & \textbf{31.2035} & \textbf{0.8492} \\
\bottomrule
\end{tabular}}
    \vspace{-0.1in}
\end{center}
\end{table}

\subsection{Deformable Convolution}\label{sec:exp_deform_conv}
After deciding on the final end-to-end architecture, we further introduce deformable convolution to adaptively learn the receptive field. We compare the performance of end-to-end architectures with and without deformable convolution, and visualize the sampling location of normal convolution and deformable convolution.

\begin{figure}[t]
\begin{center}
  \includegraphics[width=\linewidth]{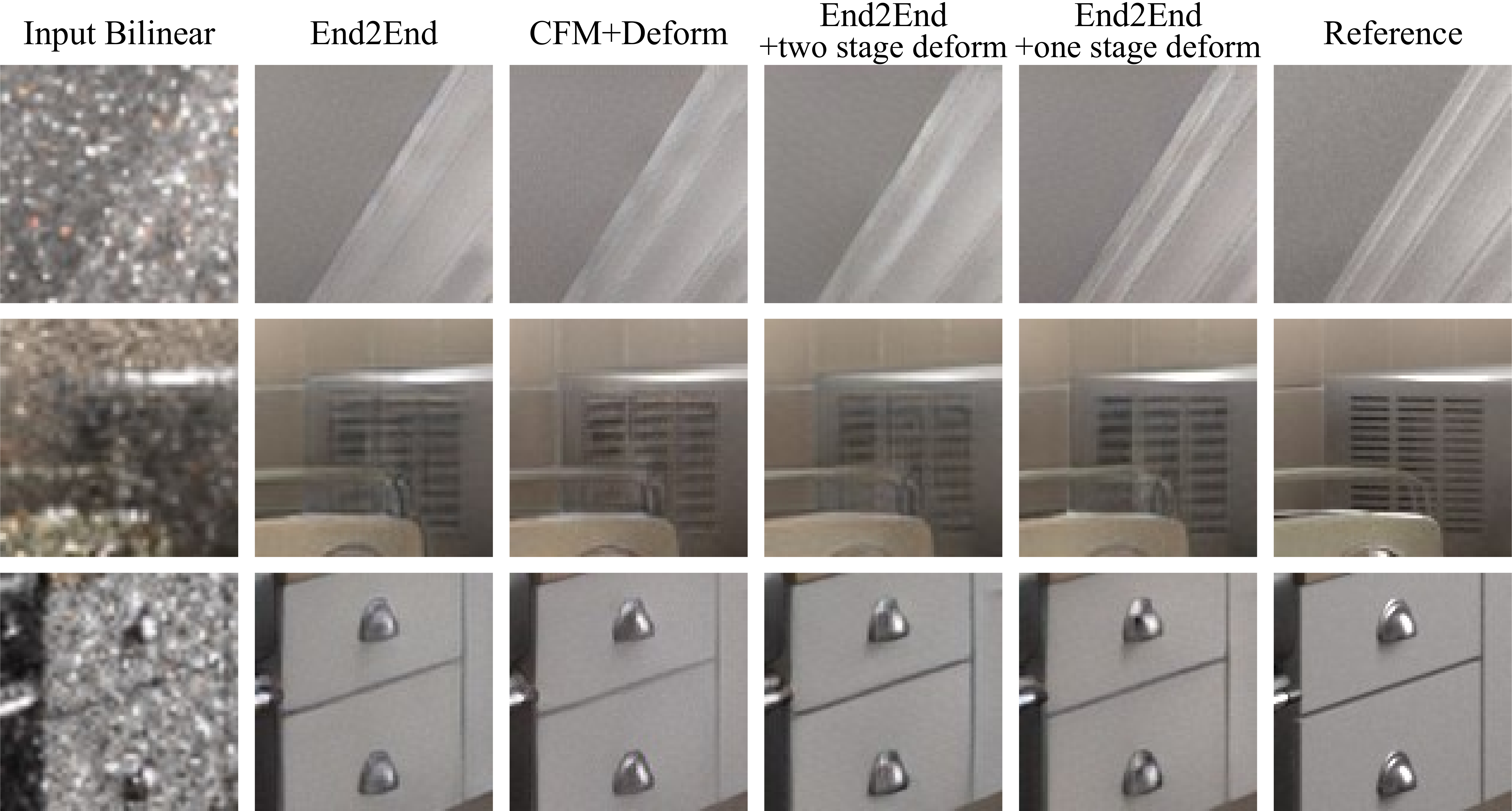}
\end{center}
    \vspace{-0.15in}
  \caption{Comparison between models w/o and w/ deformable convolution and different ways of introducing it. Adding deformable convolution only in denoising stage has the best performance, especially achieves clear edges and outlines.}
    \vspace{-0.1in}
\label{fig:deform}
\end{figure}

We explore three ways of introducing deformable convolution for the SRD task. The first one is simply replacing the second convolution layer in each residual block of CFM~\cite{xu2019adversarial} with deformable convolution, denoted as \textit{CFM+deform}. 
The second trial is integrating deformable convolution into both SR and denoising stages, denoted as \textit{End2End+two stage deform}. The third way is only using deformable convolution in the denoising stage, denoted as \textit{End2End+one stage deform}. Note that the \textit{End2End} model refers to the end-to-end model (SR->denoising) with IPT. The results are shown in Table~\ref{tab:deform}. We can see that even being able to adaptively learn the receptive field, the one-stage model still performs worse than two-stage models. For \textit{End2End+two stage deform}, the performance is weaker than \textit{End2End+one stage deform}. This is because the original 3$\times$3 convolution is small enough for the SR task and using deformable only adds to training complexity, which hinders the final performance. The results in Fig.~\ref{fig:deform} show the visual difference. Overall, adding deformable convolution leads to sharper edges. Compared with \textit{End2End+one stage deform}, the other methods produce more blurry results with fewer details.

\begin{table}
    \caption{\label{tab:deform}Comparison among three ways of introducing deformable convolution for the SRD task. \textit{CFM+deform} means using deformable convolution in a one-stage CFM model. \textit{End2End+x stage} means using deformable convolution in both or only the second stages for the end-to-end model. \textit{End2End} model refers to the end-to-end model (SR->denoising) with IPT.}
    \vspace{-0.15in}
\begin{center}
\begin{tabular}{l|ccc}
\toprule
Model & relMSE$\downarrow$ & PSNR$\uparrow$ & SSIM$\uparrow$ \\
\hline
Input Bilinear & 0.0648 & 21.4412 & 0.4096 \\
CFM+deform & 0.0070 & 31.3201 & 0.8540 \\
End2End+two stage deform & 0.0060 & 31.6700 & 0.8572 \\
End2End+one stage deform & \textbf{0.0055} & \textbf{32.1106} & \textbf{0.8608} \\
\bottomrule
\end{tabular}
    \vspace{-0.1in}
\end{center}
\end{table}

\begin{figure}[t]
\begin{center}
   \includegraphics[width=\linewidth]{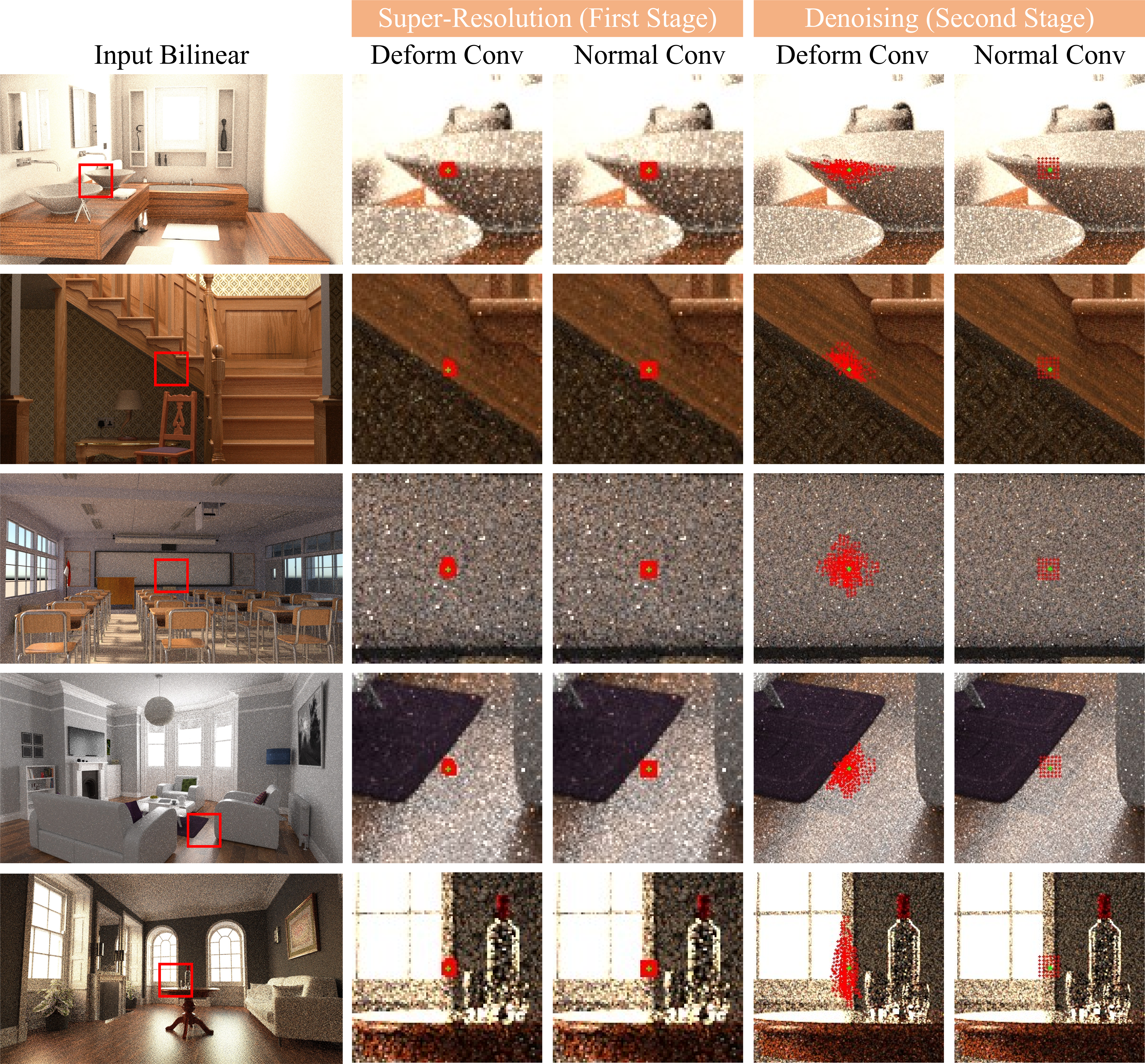}
\end{center}
    \vspace{-0.15in}
  \caption{Visualization of the sampling locations of deformable and normal convolutions in each stage of the end-to-end network. The green point refers to the center pixel and red points refer to sampling locations.}
    \vspace{-0.1in}
\label{fig:deform_vis}
\end{figure}

To show the difference of sampling region for the SR or denoising task, we visualize the results of \textit{End2End+two stage deform}, which is shown in Fig~\ref{fig:deform_vis}.  We choose five typical situations to compare the difference between normal and deformable convolutions, including edges, strip regions and blank areas. Normal convolution always has a fixed sampling region regardless of the semantic information, while deformable convolution varies according to different situations. We can observe the learned sampling region shrinks in the SR stage, while expanding in the denoising stage. This observation is consistent with the receptive field exploration experiment in Sec.~\ref{sec:exp_rf}. Visually, compared with normal convolution, we can also see that using deformable convolution in the SR stage brings little difference. Thus, we just employ normal convolution in the SR stage to avoid introducing unnecessary computational cost.

\subsection{Comparison with State-of-the-art}\label{sec:exp_sota}
In order to compare our proposed end-to-end architecture with other methods, we adapt state-of-the-art super-resolution and denoising methods into the SRD setting.

Overall, our method outperforms other adapted methods both quantitatively and qualitatively. Table~\ref{tab:sota} shows the quantitative scores of each method using 8spp 540p images as inputs. 

\begin{table}
    \caption{\label{tab:sota}Comparison with adapted state-of-the-art methods. \textit{Input Bilinear} refers to resizing input data with the bilinear upsampling strategy.}
    \vspace{-0.15in}
\begin{center}
\resizebox{\linewidth}{!}{
\begin{tabular}{l|c|ccc}
\toprule
Model & FLOPS/GMac & relMSE$\downarrow$ & PSNR$\uparrow$ & SSIM$\uparrow$ \\
\hline
Input Bilinear & - & 0.0648 & 21.4412 & 0.4096 \\
EDSR~\cite{lim2017enhanced} & 258.10 & 0.0094 & 30.1499 & 0.8386 \\
RAE~\cite{chaitanya2017interactive} & 251.81 & 0.0104 & 30.0046 & 0.8354 \\
KPCN~\cite{bako2017kernel} & 269.42 & 0.0120 & 29.0795 & 0.8219 \\
CFM~\cite{xu2019adversarial} & 247.97 & 0.0076 & 31.1201 & 0.8436 \\
Ours & 258.29 & \textbf{0.0055} & \textbf{32.1106} & \textbf{0.8608} \\
\bottomrule
\end{tabular}}
    \vspace{-0.1in}
\end{center}
\end{table}

\begin{figure*}[t]
\begin{center}
   \includegraphics[width=\linewidth]{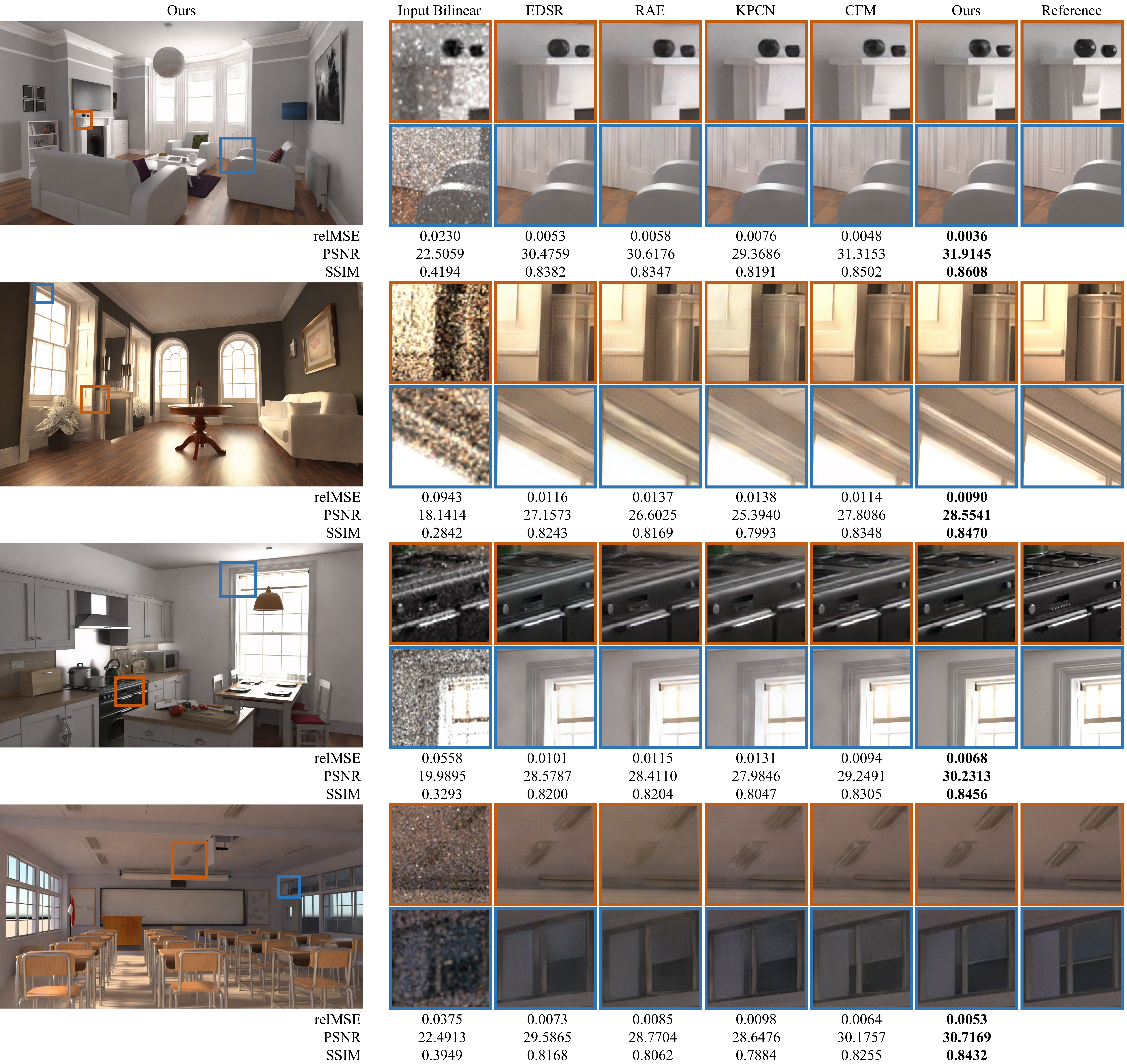}
\end{center}
    \vspace{-0.15in}
  \caption{Comparison with the adapted state-of-the-art methods. With the benefit of deformable convolution, our method achieves better performance on structures.}
    \vspace{-0.1in}
\label{fig:sota}
\end{figure*}

Fig.~\ref{fig:sota} shows the qualitative results. We can see that our method contains more and clearer details, while others like RAE and KPCN overblurs them. Our method outputs smoother results while EDSR and CFM usually generate obvious color stains, such as the second result in the first scene. Moreover, with the benefit of deformable convolution, our method achieves better results on structure content, such as better recovering straight lines around the windows.

\subsection{Time Evaluation}\label{sec:time_eval}
We evaluate the running time of Monte Carlo rendering and post-processing methods to prove the motivation of SRD task. Our path tracing results are based on Mitsuba renderer running on a 12-core CPU machine. Generating one frame of 1080p 8spp rendering requires 19.33s while a 540p 8spp rendering requires only 4.45s, which is much closer to 1080p 1spp rendering (2.02s). And the deep learning based post-processing method generates the output within 1s on a GTX 1080Ti GPU, thus the total processing time required by a noise-free image is reduced to one-fifth by the SRD setup. 

\section{Analysis}

\subsection{Effectiveness of Information Propagation Techniques}\label{sec:exp_info_reserve}
After deciding the \textit{End2End} baseline model, we further explore techniques to better propagate information in an end-to-end manner. The ablation study in Table~\ref{tab:info_reserve} shows the improvement brought by each technique. First, we enlarge the output channel of the first stage from three to the same number as that of the former layer, denoted as \textit{channel aug}. Then we add a skip connection between the two stages, denoted as \textit{skip}. Finally, we change the way of utilizing gBuffer data from simply concatenating to a conditioned feature module (CFM).

\begin{table}
    \caption{\label{tab:info_reserve}Ablation study of different information propagation techniques.}
    \vspace{-0.15in}
\begin{center}
\resizebox{\linewidth}{!}{
\begin{tabular}{l|ccc}
\toprule
Model & relMSE$\downarrow$ & PSNR$\uparrow$ & SSIM$\uparrow$ \\
\hline
Simple End2End (SR->denoising) & 0.0094 & 30.4851 & 0.8432 \\
+channel aug & 0.0094 & 30.6855 & 0.8464 \\
+channel aug+skip & \textbf{0.0075} & 30.9012 & 0.8460 \\
+channel aug+skip+CFM & 0.0076 & \textbf{31.2035} & \textbf{0.8492} \\
\bottomrule
\end{tabular}}
    \vspace{-0.1in}
\end{center}
\end{table}

\begin{figure*}[t]
\begin{center}
   \includegraphics[width=\linewidth]{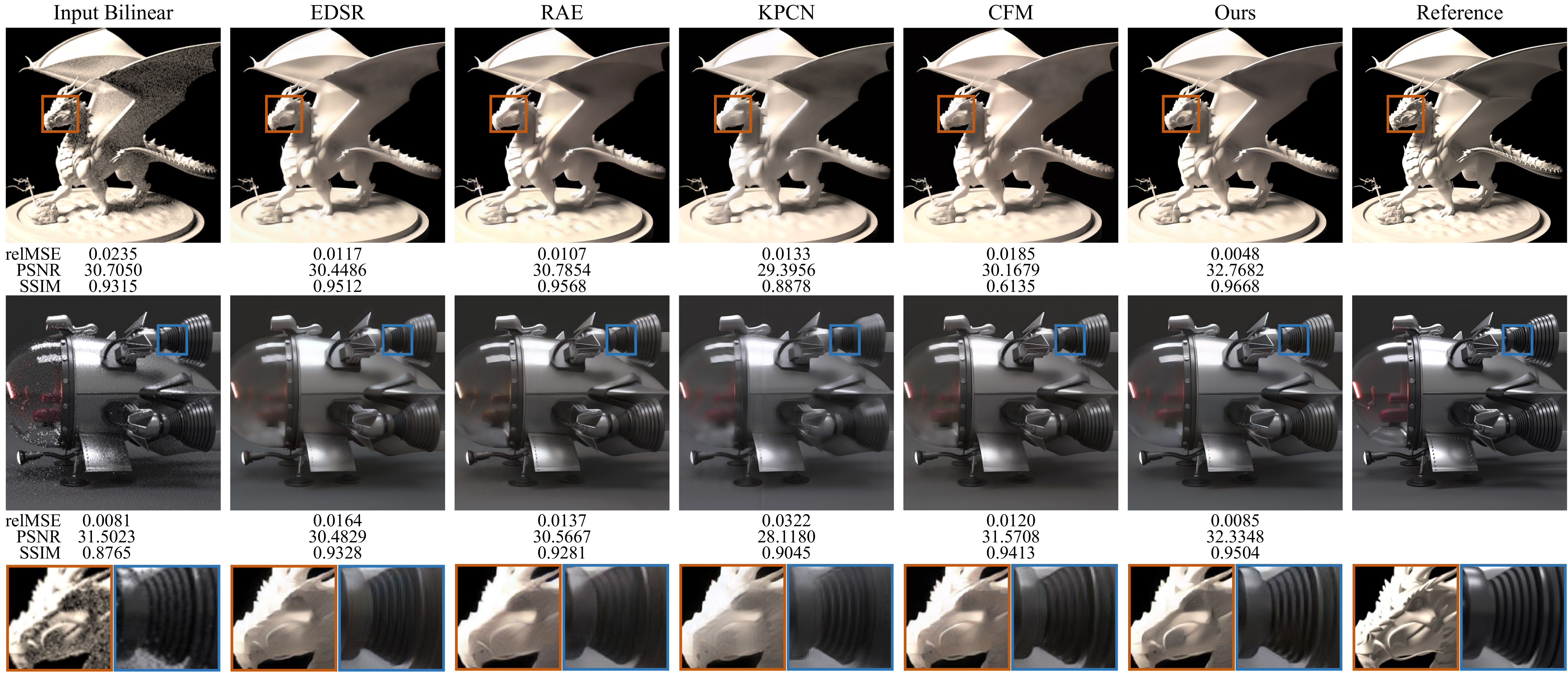}
\end{center}
    \vspace{-0.15in}
  \caption{Evaluation on generalization ability. The facial structures and the shadow effect of the dragon are better kept by our method. And our method achieves cleaner edges while others contain a blur effect. Moreover, for the glass mask, with little support from gBuffer data, the results of our method are more reasonable and closer to the ground truth.}
    \vspace{-0.1in}
\label{fig:general}
\end{figure*}

\subsection{Number of Deform. Blocks}\label{sec:deform_num}
We compare the difference brought by the number of deformable blocks. As stated in Sec.~\ref{sec:deform_conv}, we replace the normal convolution in the second layer of each block with deformable convolution to produce a deformable block. Here, we conduct an ablation study by changing the last $n$ blocks into the deformable block. The results in Table~\ref{tab:deform_num} shows that using deformable convolution always brings performance improvement. When the number of deformable blocks reaches three, we achieve the best performance. Too many deformable convolution blocks oppositely reduce the improvement because it becomes more and more difficult for the network to converge to an optimal result, which is consistent with the trade-off analysis in Sec.~\ref{sec:deform_conv}.


\begin{table}
    \caption{\label{tab:deform_num}Models with different number of deformable blocks in the second stage. The block name \textit{ux} is corresponding to the x-th upsampling block.}
    \vspace{-0.15in}
\begin{center}
\begin{tabular}{l|ccc}
\toprule
\#Deform Block & relMSE$\downarrow$ & PSNR$\uparrow$ & SSIM$\uparrow$ \\
\hline
0 & 0.0076 & 31.2035 & 0.8492 \\
u1 & 0.0063 & 31.6034 & 0.8552 \\
u1+u2 & 0.0062 & 31.6079 & 0.8554 \\
u1+u2+u3 & \textbf{0.0055} & \textbf{32.1106} & \textbf{0.8608} \\
u1+u2+u3+u4 & 0.0061 & 31.9046 & 0.8586 \\
u1+u2+u3+u4+u5 & 0.0061 & 31.7493 & 0.8576 \\
\bottomrule
\end{tabular}
    \vspace{-0.1in}
\end{center}
\end{table}

\subsection{Effectiveness of Perceptual Loss}\label{sec:exp_preceptual}
\begin{figure}[t]
\begin{center}
   \includegraphics[width=\linewidth]{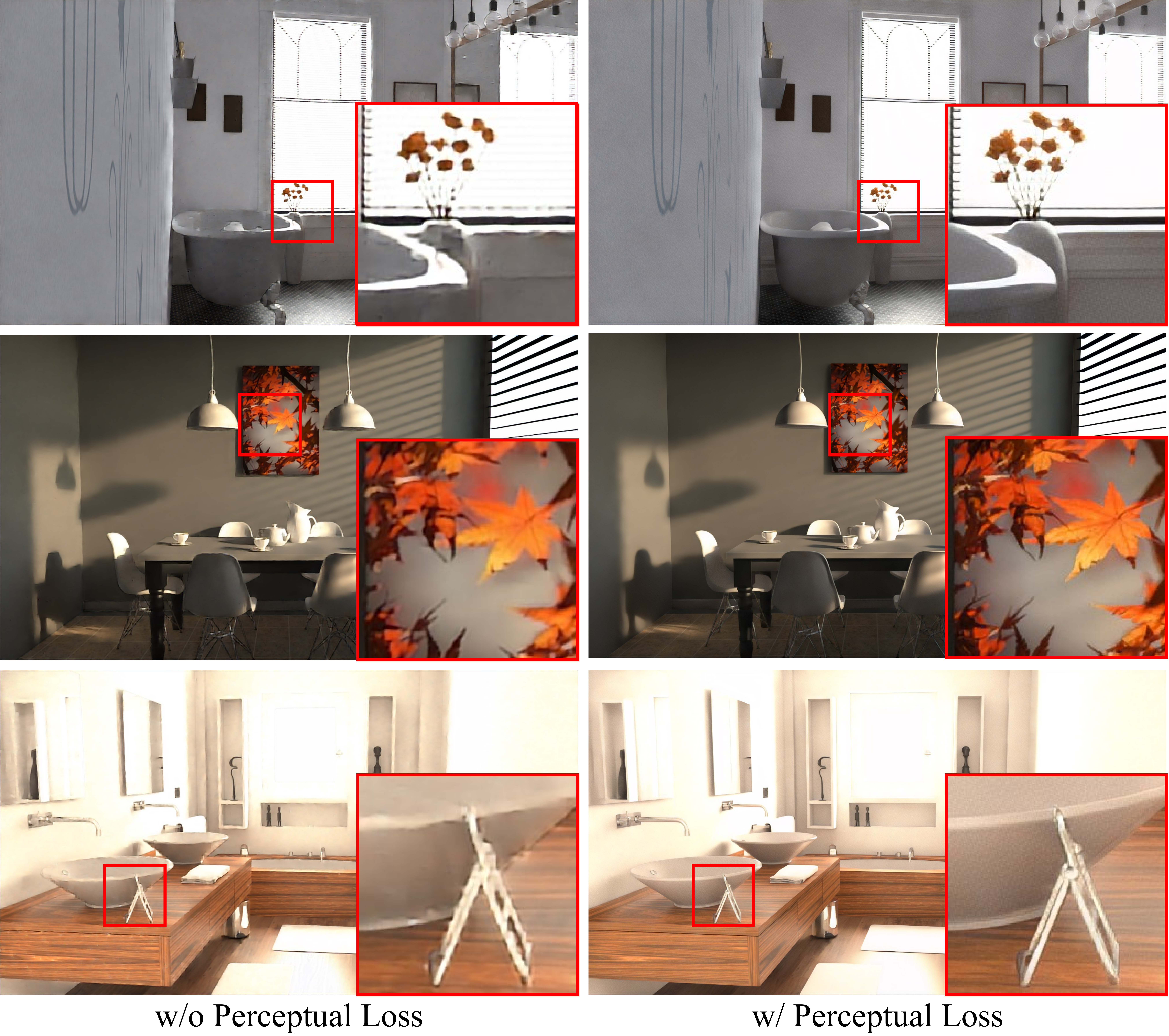}
\end{center}
    \vspace{-0.15in}
    \caption{\label{fig:perceptual}Comparison between models w/ and w/o perceptual loss. Adding perceptual loss helps to recover better details.}
    \vspace{-0.1in}
\end{figure}

To illustrate the superiority of perceptual loss in reserving details and structural information, we compare the models with and without perceptual loss, as shown in Fig.~\ref{fig:perceptual}. Perceptual loss helps the bathtub and mirror to retain sharper edges, while the flowers and leaves contain clearer details.

\subsection{Generalization Ability}
We evaluate the model generalization ability by testing its performance on scenes outside the training set. 
The results are shown in Fig.~\ref{fig:general}. As for the first case, the facial structures of the dragon are mostly lost by other state-of-the-art methods, while our method keeps more details and recovers better shadow effect. In the second example, our method achieves cleaner edges while others contain a blur effect. Moreover, for the glass mask, where gBuffer offers little auxiliary support, the results of our method are more reasonable and closer to the ground truth.

\subsection{Discussion}
In the process of training model, we found that the diversity of the training data affects the quality of recovered color. In detail, training with few types of scenes causes color fading, especially the bright colors. So in our experiments, we pick ten scenes with various color types to prevent this problem. The current results look good but using more scenes with higher diversity in color can lead to better color recovery.

We notice that other denoising methods, such as KPCN and CFM, process specular and diffuse as two individual branches. We did not apply this operation because it is out of the scope of our paper, though it may be effective in improving performance.


\section{Conclusions and Future Work}

We present the first deep neural network conducting super-resolution and denoising simultaneously in the post-processing stage for accelerating Monte Carlo rendering. We explore the network characteristics requirement of the novel SRD task, especially the impact of the receptive field. We introduce deformable convolution for adaptively adjust the receptive field. We show that our method has stronger ability of generating smoother visual effect as well as keeping more structural details.

In the future, we would like to collect a larger amount of data for training and testing to explore more about the generalization ability of our model. In addition, it is interesting to study combining more post-processing techniques in a neural network besides denoising and super-resolution, such as tone mapping and bloom. The more a single neural network can accomplish in an end-to-end manner, the more computational cost we can save. Finally, although our current model has the ability to maintain temporal consistency with the help of RNN mechanism and the temporal loss, it is also promising to explicitly utilize more history frames to achieve better results as in ~\cite{xiao2020neural}.

%
%
%
%

\bibliographystyle{ACM-Reference-Format}
\bibliography{main}

\end{document}